\documentclass[pdflatex,sn-nature]{sn-jnl}

\usepackage{graphicx}
\usepackage{siunitx}
\usepackage{multirow}
\usepackage{amsmath,amssymb,amsfonts}
\usepackage{amsthm}
\usepackage{mathrsfs}
\usepackage[title]{appendix}
\usepackage{xcolor}
\usepackage{textcomp}
\usepackage{manyfoot}
\usepackage{booktabs}
\usepackage{algorithm}
\usepackage{algorithmicx}
\usepackage{algpseudocode}
\usepackage{listings}
\usepackage{url}
\usepackage{longtable}
\usepackage[left]{lineno}

\theoremstyle{thmstyleone}

\theoremstyle{thmstyletwo}

\theoremstyle{thmstylethree}

\begin{document}

\title[M3OS]{M3OS: A Monte Carlo Graph Search–Orchestrated Multi-Agent LLM System for Evidence-Traced Molecular Optimization}

\author[1,2]{\fnm{Junjie} \sur{Wang}}
\equalcont{These authors contributed equally to this work.}

\author[1]{\fnm{Yaowei} \sur{Jin}}
\equalcont{These authors contributed equally to this work.}

\author[1]{\fnm{Ruohui} \sur{Tang}}
\equalcont{These authors contributed equally to this work.}

\author[1]{\fnm{Guonan} \sur{Cui}}

\author[1]{\fnm{Haojie} \sur{Wang}}

\author[1]{\fnm{Penglei} \sur{Wang}}

\author[1]{\fnm{Dingyan} \sur{Wang}}

\author[1]{\fnm{Duo} \sur{An}}

\author[1,3]{\fnm{Shuangjia} \sur{Zheng}}

\author*[1]{\fnm{Qian} \sur{Shi}}\email{shiqian@lglab.ac.cn}

\affil[1]{\orgname{Lingang Laboratory}, \orgaddress{\city{Shanghai}, \postcode{200031}, \country{China}}}

\affil[2]{\orgdiv{School of Information Science and Technology}, \orgname{ShanghaiTech University}, \orgaddress{\city{Shanghai}, \postcode{201210}, \country{China}}}

\affil[3]{\orgdiv{Global Institute of Future Technology}, \orgname{Shanghai Jiao Tong University}, \orgaddress{\city{Shanghai}, \postcode{200240}, \country{China}}}



\abstract{Small-molecule optimization integrates medicinal-chemistry reasoning and computational evidence through iterative, multi-objective decisions. When large language models (LLMs) reason over optimization histories stored primarily in conversational context, they must recover candidate identities, prior evaluations, and task constraints to guide subsequent decisions. We present M3OS, a multi-agent LLM system that decouples molecular-design reasoning from optimization-state management through Monte Carlo graph search. A persistent graph links evaluated candidates, parent–child transformations and evaluation evidence, while rewards and visit statistics guide LLM-assisted parent selection. Two branches combine tool-driven candidate generation with knowledge- and case-guided medicinal-chemistry editing. An execution harness controls graph updates through structured output extraction, molecular validation and task-bound evaluation. Agents receive role-specific contexts, while the graph preserves optimization trajectories beyond their active contexts. Across three molecular optimization benchmarks, M3OS achieves higher success rates than baselines, supporting the integration of persistent search state, specialized agents and controlled execution for multi-constraint optimization.}


\maketitle

\section{Introduction}\label{sec:introduction}

Early-stage drug discovery requires the simultaneous optimization of multiple, often competing objectives that collectively determine the therapeutic potential and developability of molecular candidates \cite{zhang2024deep}. In practice, medicinal chemists progressively refine candidate molecules by preserving favorable structural motifs and pharmacophores, mitigating physicochemical liabilities, and balancing bioactivity against absorption, distribution, metabolism, excretion, and toxicity (ADMET) properties \cite{d2012multi, racz2025changing}. Medicinal-chemistry knowledge and accumulated experimental evidence guide both the choice of structural modifications and the selection of candidates for further optimization.

Many existing studies formulate molecular optimization as a combinatorial optimization problem and employ techniques such as Bayesian optimization \cite{korovina2020chembo}, genetic algorithms \cite{jensen2019graph}, Monte Carlo tree search \cite{zhang2025molecular}, and reinforcement learning \cite{zhou2019optimization, you2018graph}. Generative approaches, including diffusion models and Bayesian Flow Networks, further expand the methods available for proposing molecular structures \cite{dorna2024tagmol, qiu2024empower}.

Large language models (LLMs) provide a means of combining natural-language objectives with chemical knowledge and tool-assisted reasoning for molecular design \cite{wang2026survey}. Agent frameworks extend this capability through planning, memory and interaction with external tools, allowing computational feedback to inform successive decisions \cite{wang2023voyager}.

ChemLLM \cite{zhang2024chemllm}, LlaSMol \cite{yu2024llasmol} and MolX \cite{le2024molx} develop chemical language and representation capabilities through domain-specific training and multimodal learning. For molecular optimization, MOLLEO integrates chemistry-aware LLMs into evolutionary mutation and crossover \cite{wang2025efficient}, while MT-Mol combines specialized agents with tool-guided reasoning and iterative molecular refinement \cite{kim2025mt}. DrugAgent \cite{inoue2025drugagent}, PharmAgents \cite{gao2025pharmagents} and DrugPilot \cite{li2025drugpilot} extend agent-based methods to broader drug-discovery workflows.

However, iterative molecular optimization requires more than proposing a plausible edit at each step. Each new decision must remain connected to previously evaluated structures, their transformation history and the task requirements. When this information is carried primarily in conversational context, agents must repeatedly recover it from an evolving text record, creating opportunities for omitted evidence, incorrect candidate–result associations or inconsistent constraint interpretation. The challenge is therefore to maintain a persistent, candidate-linked optimization state while presenting each agent with the information needed for its current decision, without making each design agent responsible for reconstructing the overall optimization history.

M3OS addresses this challenge by decoupling molecular-design reasoning from optimization-state management through a persistent molecular search graph (Fig.~\ref{fig:framework}a). Canonical molecular identities define nodes that retain generation provenance, evaluation evidence and search statistics, while edges explicitly record parent–child transformations. Rediscovered molecules reuse existing nodes while preserving alternative incoming transformations. This graph serves as the operational state for subsequent decisions, rather than only a retrospective record. Monte Carlo graph search (MCGS) \cite{czech2020monte} uses accumulated rewards and visit statistics to shortlist expansion parents, and an LLM selects among them using task context and node-level evidence. Previously evaluated intermediates remain available for further expansion, connecting the local decision of which molecular edit to propose with the longer-term decision of which optimization trajectory to continue.

Within each expansion, two specialized agents contribute distinct proposal mechanisms to this shared state. The Creative Molecule Explorer generates and screens candidate pools using molecular-design tools, whereas the Rational Medicinal Designer develops targeted edits supported by medicinal-chemistry knowledge and scaffold-matched optimization cases. Both receive the same selected parent, shared molecular analysis and completed-round screening feedback, including remaining target gaps and failure reasons, while retaining separate role-specific reasoning memories. The MedChem Retrieval Agent supplies concise, source-linked guidance while keeping detailed retrieval context local. A common Critic evaluates both branches against the same task contract. This organization lets each design agent focus on its proposal mechanism while sharing the evaluated outcomes needed to continue optimization across rounds and branches.

An execution harness connects this open-ended reasoning process to controlled updates of the shared state. The Chief Architect formalizes objectives and constraints as a task contract, while procedural instructions and role-specific tool restrictions guide execution. Agents provide natural-language design and evaluation rationales; an extraction-only Auditor converts their finalized outputs into predefined records using only explicitly stated molecules and evidence. Molecular validation, canonical identity checks and structured-evaluation requirements govern graph admission, without excluding valid, evaluated intermediates solely because they do not yet satisfy the task. This separates the flexibility of free-text reasoning from the consistency requirements of persistent records: LLMs remain involved in parent selection, molecular design and comparative assessment, while the harness governs execution boundaries and state updates. Recorded molecular transformations, rationales and evaluations remain linked for subsequent decisions and inspection by an on-demand Reporter.

In summary, our contributions are as follows:
\begin{itemize}
    \item \textbf{Persistent-state graph-search orchestration.}
    We introduce M3OS, an MCGS-orchestrated framework that decouples molecular-design reasoning from optimization-state management. A persistent graph explicitly links molecular identities, transformation trajectories and evaluation evidence, retaining reusable history beyond individual agents' conversational contexts. Accumulated rewards and visit statistics guide LLM-assisted parent selection, enabling subsequent rounds to revisit and extend previously evaluated intermediates.

    \item \textbf{Specialized collaboration with controlled execution.}
    We develop an execution harness that coordinates tool-driven candidate generation and knowledge- and case-guided medicinal-chemistry editing under a common task contract and shared evaluation feedback. Role-specific contexts support complementary design strategies, while free-form design rationales are retained alongside structured molecular and evaluation records. Structured extraction and molecular validation govern graph updates, coupling flexible agent reasoning with controlled execution and traceable optimization.

    \item \textbf{Improved performance across three optimization benchmarks.}
    Experiments on MolOpt-120, MuMOInstruct-100 and SMDD-Bench-93 demonstrate higher mean candidate-pool success in property optimization and higher final-solution success in activity-constrained lead optimization than Codex and Claude Code using the same LLM backbone and molecular-tool access.
\end{itemize}

\section{Results}\label{sec:results}

\subsection{Overview of M3OS}\label{sec:framework_overview}

\begin{figure}[!ht]
\centering
\includegraphics[width=\linewidth]{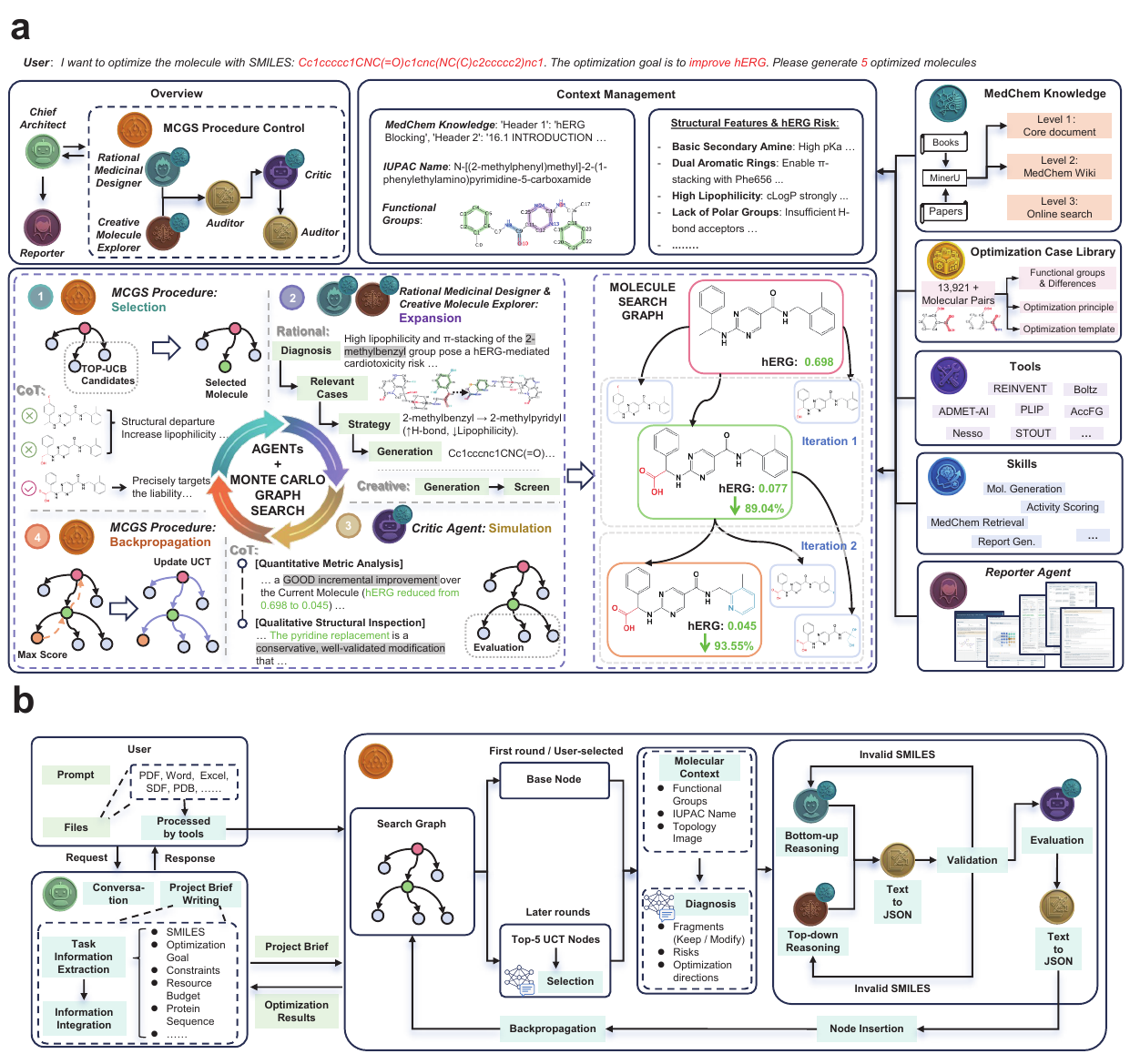}
\caption{M3OS architecture and optimization loop. \textbf{a}, Specialized agents share a persistent molecular graph that links candidates, structural transformations and evaluation evidence. \textbf{b}, Node selection, parallel Creative and Rational generation, molecular validation, common Critic evaluation and reward updates connect successive optimization rounds. Knowledge resources, tools and procedural skills support these operations.}
\label{fig:framework}
\end{figure}

Each optimization round expands a selected molecular parent and incorporates the evaluated proposals into the persistent search graph (Fig.~\ref{fig:framework}). The Chief Architect translates the starting molecule, objectives and constraints into a task brief. MCGS uses accumulated evaluations and visit statistics to rank expansion candidates, and an LLM selects a parent from the resulting shortlist using task context.

The selected parent and previous screening feedback are provided to both generation branches. The Creative Molecule Explorer builds and screens proposals with molecular-generation tools. The Rational Medicinal Designer develops molecular edits using medicinal-chemistry guidance and scaffold-matched transformation examples. Proposals from both branches undergo molecular validation, deduplication and Critic evaluation against the task contract.

The execution harness converts finalized proposals and evaluations into structured graph records. Candidate identity, parent transformations, tool results and search statistics remain available for subsequent selection and expansion. 

For post-search inspection, the Reporter converts a graph snapshot into a molecular decision report. Before delivery, deterministic checks verify one graph node and one detail record per molecule, one rendered edge per valid graph relationship, consistent molecule identifiers, complete SMILES and direct-parent comparisons (Appendix~\ref{app:reporting}).

\subsection{Benchmark Datasets and Subset Construction}

We evaluated M3OS on three benchmark subsets covering property optimization and activity-constrained lead optimization. MolOpt-120  contains 120 starting molecules across eight single- or double-property tasks \cite{ye2025drugassist}, while MuMOInstruct-100 contains 100 records across five tasks involving three or four properties \cite{dey2025gellm3o}. SMDD-Bench-93 comprises 93 lead-optimization tasks requiring improved predicted target activity while satisfying physicochemical, ADMET and structural constraints \cite{han2026smdd}. Subset construction and adaptations to the original evaluation are detailed in Appendix~\ref{app:exp}.

\subsection{Baselines}

We compare M3OS with chemistry-specific language models, direct general-purpose language models, and general-purpose coding agents. The chemistry-specific baselines include: (1) ChemLLM-7B-Chat-1.5-SFT \cite{zhang2024chemllm}, a conversational language model instruction-tuned for chemical knowledge and reasoning; (2) DrugAssist-7B \cite{ye2025drugassist}, an interactive Llama-2 model fine-tuned on MolOpt-Instructions for single- and multi-property molecule optimization; (3) GeLLM$^{3}$O-P6-Mistral \cite{dey2025gellm3o}, a Mistral-based model instruction-tuned on MuMOInstruct to generalize across combinations of molecular properties; (4) LlaSMol-CodeLlama-7B \cite{yu2024llasmol}, which is instruction-tuned on the multi-task SMolInstruct corpus for chemical generation, prediction, conversion and reaction tasks; (5) mCLM-3B \cite{edwards2026mclm}, a modular chemical language model that represents molecules using functional and synthesis-oriented building blocks. 
The direct general-purpose language-model baselines include: (1) GPT-5.5; (2) Claude Opus 4.6; (3) Gemini 3.5 Flash; (4) Kimi-K3 Low (Reasoning effort=Low); (5) Kimi-K3 High (Reasoning effort=High). 
The general-purpose coding-agent baselines are (1) Codex and (2) Claude Code. Both use Kimi-K3 Low as their underlying LLM and have access to the full MCP tool set.
Candidate-pool construction and benchmark-specific evaluation are described in Section~\ref{sec:metrics}.

\subsection{Metrics}\label{sec:metrics}

For MolOpt-120 and MuMOInstruct-100, we evaluate candidate pools produced under method-specific generation and selection protocols. Fixed-pool baselines generate 20 candidates per record, and coding agents return up to 20. M3OS pools contain up to 20 non-root, canonical-unique molecules selected post hoc from each cumulative search graph using system-ranked results for benchmark success, directional property gains and similarity. Repeated baseline outputs are counted separately. RDKit \cite{bento2020open} evaluates molecular validity and physicochemical descriptors; similarity $T(x,y)$ is the Tanimoto coefficient \cite{bajusz2015tanimoto} of 2,048-bit Morgan fingerprints \cite{rogers2010extended} with radius 2 and no chirality encoding. ADMET-AI \cite{swanson2024admetai} supplies ADMET predictions. 

\paragraph{MolOpt-120.}
Loose success requires every requested property to improve strictly relative to the initial molecule. Strict success additionally requires all applicable thresholds: increases $>0.1$ for QED and BBBP, increases of at least two for Hydrogen Bond Acceptor (HBA) and Hydrogen Bond Donor (HBD), a solubility increase strictly between 0.5 and 1.5, and an hERG decrease $>0.1$. For QED or BBBP with an initial value above 0.9, and for hERG with an initial value below 0.1, strict directional improvement satisfies the corresponding property criterion. For task $t$ with 15 records,
\begin{equation}
\mathrm{SR}_{t}^{(q)}
=\frac{1}{15}\sum_{i\in\mathcal{I}_t}\frac{1}{n_i}
\sum_{j=1}^{n_i}s_{ij}^{(q)},
\qquad q\in\{\mathrm{loose},\mathrm{strict}\},
\label{eq:molopt_metrics}
\end{equation}
where $\mathcal{I}_t$ indexes task records, $n_i$ is the candidate-pool size, and $s_{ij}^{(q)}$ indicates candidate success. Invalid candidates count as failures and remain in the denominator. Validity uses the same averaging with a validity indicator. Overall rates are macro-averaged across the eight tasks.

\paragraph{MuMOInstruct-100.}
For record $i$, let $x_i$ be the initial molecule, $y_{ij}$ a candidate, $\mathcal{V}_i$ the valid-candidate indices and $\mathcal{P}_i$ the requested properties. SR is the fraction of valid candidates that strictly improve every requested property. Sim is the mean $T(x_i,y_{ij})$ over all valid candidates. Directional relative improvement is
\begin{equation}
\mathrm{RI}_i=
\frac{1}{|\mathcal{V}_i|\,|\mathcal{P}_i|}
\sum_{j\in\mathcal{V}_i}\sum_{p\in\mathcal{P}_i}
\frac{d_p\bigl(f_p(y_{ij})-b_{ip}\bigr)}{|b_{ip}|},
\label{eq:mumo_record_metrics}
\end{equation}
where $f_p$ is the candidate property value, $b_{ip}$ is its nonzero source baseline, and $d_p=+1$ for BBBP, HIA, pLogP and QED, and $-1$ for mutagenicity. Similarity and RI include valid candidates that fail joint success. Each metric is averaged over records with valid candidates and then across the five tasks. Reported $\mathrm{SR}\times\mathrm{Sim}$ multiplies the corresponding values.

\paragraph{SMDD-Bench-93.}
A candidate succeeds only if it satisfies all benchmark physicochemical, structural, ADMET optimization and hold constraints, including similarity $\geq0.7$ to the reference ligand. Nesso replaces Boltz for activity evaluation across all methods due to its faster speed and more accurate predictions, requiring binding probability $>0.7$ and affinity improvement $a_i^{\mathrm{ref}}-a_i(y)\geq0.30$. Both affinity values are predicted for the same protein target, with lower values preferred. Let $P_i(y)$ indicate satisfaction of all requirements. For final output $y_i^{\mathrm{final}}$ and the set of recorded graph molecules $V_i^{\mathrm{graph}}$,
\begin{equation}
\begin{aligned}
\mathrm{SR}_{\mathrm{single}}
&=\frac{1}{93}\sum_{i=1}^{93}P_i(y_i^{\mathrm{final}}),\\
\mathrm{SR}_{\mathrm{graph\mbox{-}any}}
&=\frac{1}{93}\sum_{i=1}^{93}
\mathbf{1}\!\left[\exists y\in V_i^{\mathrm{graph}}:P_i(y)=1\right].
\end{aligned}
\label{eq:smdd_success_metrics}
\end{equation}
Graph-any evaluates the full graph without a 20-candidate cap, with each task contributing at most one success.

\subsection{Performance Comparison}

 R1--R3 denote successive M3OS search rounds; MolOpt-120 and MuMOInstruct-100 report system-ranked cumulative candidate pools, whereas SMDD-Bench-93 separately evaluates final single-solution and graph-any success (Section~\ref{sec:metrics}).

\subsubsection{Threshold-based property improvement on MolOpt-120.}

The system-ranked M3OS R3 pools achieve the highest loose success rates on seven of the eight MolOpt-120 tasks and the highest strict success rates on five tasks (Table~\ref{tab}). Mean strict success across the eight tasks increases from approximately 0.47 at R1 to 0.70 at R3; the submitted Claude Code and Codex pools achieve approximately 0.61 and 0.60, respectively.

\subsubsection{Joint optimization of multiple properties on MuMOInstruct-100.}


On MuMOInstruct-100, the system-ranked M3OS R3 pools achieve the highest joint success rate on three of the five tasks, with a mean joint success rate of approximately 0.85 across tasks; the submitted Claude Code and Codex pools achieve approximately 0.80 and 0.74, respectively (Table~\ref{tab:mumo-baseline-results}). Both joint success rate and relative improvement increase from R1 to R3 across all five tasks under cumulative pool evaluation. On BMPQ, for example, joint success rises from 0.76 to 0.83, while mean similarity decreases from 0.46 to 0.34.

\subsubsection{Activity-constrained lead optimization.}
With Kimi-K3 Low as the shared LLM backbone, M3OS achieves the highest final single-solution success rate on SMDD-Bench-93, reaching 0.59 compared with 0.48 for Codex and 0.42 for Claude Code (Table~\ref{tab:smdd_bench93_success}).  Graph-any success rates are 0.61, 0.57 and 0.47, respectively. M3OS therefore combines the highest final-solution success with the smallest difference between graph-any and final-solution success: approximately 0.02, compared with 0.09 for Codex and 0.05 for Claude Code.

\begin{table*}[t]
\caption{MolOpt-120 success rates (0--1; 15 records per task). L/S:
loose/strict success; HBA/HBD: hydrogen-bond acceptors/donors. Rates average
candidates, then records; invalid candidates count as failures. M3OS uses up
to 20 system-ranked, unique non-root candidates per cumulative graph;
critic-score ties at the cutoff are resolved by graph insertion order.
Baseline duplicates are retained. Solubility, BBBP and hERG use ADMET-AI
1.3.1. Bold/underline mark the best/second-best unrounded rates; ties share
emphasis.}
\label{tab}
\centering
\resizebox{\textwidth}{!}{%
\begin{tabular}{@{}l*{16}{r}@{}}
\toprule
& \multicolumn{2}{c}{QED}
& \multicolumn{2}{c}{HBD}
& \multicolumn{2}{c}{HBA}
& \multicolumn{2}{c}{Sol.}
& \multicolumn{2}{c}{BBBP}
& \multicolumn{2}{c}{hERG}
& \multicolumn{2}{c}{Sol.+HBA}
& \multicolumn{2}{c}{QED+BBBP} \\
\cmidrule(lr){2-3}\cmidrule(lr){4-5}\cmidrule(lr){6-7}
\cmidrule(lr){8-9}\cmidrule(lr){10-11}\cmidrule(lr){12-13}
\cmidrule(lr){14-15}\cmidrule(lr){16-17}
Model
& L$\uparrow$ & S$\uparrow$ & L$\uparrow$ & S$\uparrow$
& L$\uparrow$ & S$\uparrow$ & L$\uparrow$ & S$\uparrow$
& L$\uparrow$ & S$\uparrow$ & L$\uparrow$ & S$\uparrow$
& L$\uparrow$ & S$\uparrow$ & L$\uparrow$ & S$\uparrow$ \\
\midrule
\multicolumn{17}{c}{\textbf{Direct general-purpose language models}} \\
GPT-5.5
& 0.94 & 0.79 & 0.92 & 0.28 & 0.85 & 0.32 & 0.83 & 0.37
& 0.86 & 0.67 & 0.66 & 0.32 & 0.65 & 0.17 & 0.68 & 0.37 \\
Claude Opus 4.6
& 0.95 & 0.78 & 0.93 & 0.24 & 0.83 & 0.18 & 0.89 & 0.31
& 0.88 & 0.71 & 0.72 & 0.41 & 0.68 & 0.09 & 0.73 & 0.42 \\
Gemini 3.5 Flash
& 0.87 & 0.69 & 0.93 & 0.14 & 0.88 & 0.25 & 0.86 & 0.38
& 0.87 & 0.66 & 0.61 & 0.29 & 0.74 & 0.18 & 0.71 & 0.42 \\
Kimi-K3 Low
& 0.90 & 0.76 & 0.89 & 0.16 & 0.87 & 0.17
& 0.83 & \textbf{0.41} & 0.83 & 0.70 & 0.67 & 0.35
& 0.65 & 0.14 & 0.61 & 0.33 \\
Kimi-K3 Max
& 0.96 & 0.77 & \textbf{0.99} & 0.36 & 0.95 & 0.28
& 0.88 & \underline{0.41} & 0.91 & 0.77 & 0.78 & 0.48
& 0.72 & 0.12 & 0.71 & 0.44 \\
\addlinespace
\multicolumn{17}{c}{\textbf{General chemistry language models}} \\
ChemLLM
& 0.49 & 0.32 & 0.11 & 0.02 & 0.29 & 0.08 & 0.48 & 0.25
& 0.24 & 0.15 & 0.43 & 0.34 & 0.23 & 0.02 & 0.32 & 0.17 \\
LlaSMol
& 0.68 & 0.33 & 0.12 & 0.00 & 0.10 & 0.01 & 0.68 & 0.23
& 0.67 & 0.57 & 0.78 & 0.65 & 0.08 & 0.00 & 0.57 & 0.37 \\
mCLM
& 0.26 & 0.17 & 0.52 & 0.26 & 0.40 & 0.30 & 0.73 & 0.02
& 0.55 & 0.46 & 0.85 & 0.75 & 0.38 & 0.19 & 0.40 & 0.20 \\
\addlinespace
\multicolumn{17}{c}{\textbf{Molecular-optimization models}} \\
DrugAssist
& 0.69 & 0.69 & 0.65 & 0.14 & 0.40 & 0.07 & 0.45 & 0.09
& 0.56 & 0.34 & 0.49 & 0.44 & 0.23 & 0.08 & 0.63 & 0.34 \\
GeLLM$^{3}$O
& 0.83 & 0.51 & 0.11 & 0.00 & 0.11 & 0.02 & 0.43 & 0.24
& 0.64 & 0.48 & 0.42 & 0.19 & 0.06 & 0.00 & 0.58 & 0.37 \\
\addlinespace
\multicolumn{17}{c}{\textbf{Agents with full MCP tools}} \\
Kimi-K3 Low (CC)
& 0.98 & 0.87 & 0.99 & 0.56 & 0.95 & 0.53 & 0.94 & 0.33
& 0.97 & 0.93 & \underline{0.97} & 0.80 & 0.72 & 0.25
& \underline{0.88} & \underline{0.64} \\
Kimi-K3 Low (Codex)
& \underline{0.99} & \textbf{0.92} & 0.99 & 0.49
& \underline{0.98} & 0.61 & 0.93 & 0.40
& \underline{0.98} & 0.87 & 0.95 & \underline{0.81}
& 0.79 & 0.18 & 0.83 & 0.52 \\
\addlinespace
\multicolumn{17}{c}{\textbf{Ours}} \\
Kimi-K3 Low (R1)
& 0.98 & 0.84 & 0.96 & 0.24 & 0.84 & 0.26 & 0.91 & 0.30
& 0.95 & 0.85 & 0.87 & 0.59 & 0.74 & 0.21 & 0.78 & 0.50 \\
Kimi-K3 Low (R2)
& 0.97 & 0.86 & 0.98 & \underline{0.60}
& 0.94 & \underline{0.63}
& \underline{0.95} & 0.34
& 0.97 & \underline{0.93}
& 0.91 & 0.74
& \underline{0.84} & \textbf{0.32}
& 0.86 & 0.63 \\
Kimi-K3 Low (R3)
& \textbf{0.99} & \underline{0.89}
& \underline{0.99} & \textbf{0.70}
& \textbf{0.99} & \textbf{0.77}
& \textbf{1.00} & 0.33
& \textbf{1.00} & \textbf{1.00}
& \textbf{0.99} & \textbf{0.88}
& \textbf{0.89} & \underline{0.28}
& \textbf{0.94} & \textbf{0.71} \\
\bottomrule
\end{tabular}%
}
\end{table*}

\begin{table*}[t]
\caption{MuMOInstruct-100 results (20 records per task). Metrics average valid
candidates, then records with valid outputs. SR: joint success rate (0--1);
Sim.: Tanimoto similarity; RI: signed relative improvement.
Task letters B/H/M/P/Q denote
BBBP/HIA/mutagenicity/penalized logP/QED. M3OS uses up to 20 unique non-root
candidates per cumulative graph, ranked in descending order by the system
score; ties at the cutoff are resolved by graph insertion order. Baseline
duplicates are retained. Bold/underline mark the
best/second-best displayed values; ties share emphasis. Dashes indicate no
valid candidates.}
\label{tab:mumo-baseline-results}
\centering
\resizebox{\textwidth}{!}{%
\begin{tabular}{@{}l*{15}{c}@{}}
\toprule
& \multicolumn{3}{c}{BPQ}
& \multicolumn{3}{c}{MPQ}
& \multicolumn{3}{c}{BHMQ}
& \multicolumn{3}{c}{BMPQ}
& \multicolumn{3}{c}{HMPQ} \\
\cmidrule(lr){2-4}\cmidrule(lr){5-7}\cmidrule(lr){8-10}
\cmidrule(lr){11-13}\cmidrule(lr){14-16}
Model
& SR$\uparrow$ & Sim.$\uparrow$ & RI$\uparrow$
& SR$\uparrow$ & Sim.$\uparrow$ & RI$\uparrow$
& SR$\uparrow$ & Sim.$\uparrow$ & RI$\uparrow$
& SR$\uparrow$ & Sim.$\uparrow$ & RI$\uparrow$
& SR$\uparrow$ & Sim.$\uparrow$ & RI$\uparrow$ \\
\midrule

\multicolumn{16}{c}{\textbf{Direct general-purpose language models}} \\
GPT-5.5
& 0.77 & 0.46 & \textbf{1.76}
& 0.50 & 0.58 & 0.25
& 0.60 & 0.29 & 3.71
& 0.57 & 0.43 & 0.38
& 0.65 & 0.40 & 1.48 \\
Claude Opus 4.6
& 0.73 & 0.51 & \underline{1.68}
& 0.65 & 0.58 & 0.45
& 0.62 & 0.35 & 4.24
& 0.58 & 0.49 & 0.80
& 0.70 & 0.42 & 1.93 \\
Gemini 3.5 Flash
& 0.75 & 0.56 & 0.85
& 0.61 & 0.55 & 0.41
& 0.63 & 0.43 & 3.35
& 0.66 & 0.45 & 0.73
& 0.65 & 0.46 & 2.23 \\
Kimi-K3 Low
& 0.50 & 0.57 & 0.51
& 0.61 & 0.54 & 0.36
& 0.59 & 0.43 & 3.22
& 0.51 & 0.46 & 0.66
& 0.66 & 0.37 & 1.87 \\
Kimi-K3 High
& 0.76 & 0.40 & 0.72
& \underline{0.81} & 0.42 & \underline{0.68}
& 0.70 & 0.34 & \textbf{4.90}
& 0.66 & 0.34 & 1.00
& 0.72 & 0.39 & 1.92 \\

\addlinespace
\multicolumn{16}{c}{\textbf{General chemistry language models}} \\
ChemLLM
& 0.18 & \textbf{0.66} & 0.10
& 0.18 & 0.52 & -0.13
& 0.15 & 0.59 & 0.83
& 0.20 & 0.43 & 0.03
& 0.09 & 0.55 & 0.24 \\
LlaSMol
& 0.38 & 0.46 & 0.38
& 0.27 & 0.50 & 0.18
& 0.35 & 0.42 & 1.64
& 0.19 & 0.39 & 0.23
& 0.12 & 0.49 & 0.24 \\
mCLM
& 0.12 & 0.25 & 0.02
& 0.00 & 0.23 & -0.39
& 0.07 & 0.22 & 2.57
& 0.09 & 0.24 & -0.06
& 0.10 & 0.12 & 2.21 \\

\addlinespace
\multicolumn{16}{c}{\textbf{Molecular-optimization models}} \\
DrugAssist
& 0.11 & \underline{0.63} & 0.19
& 0.11 & \textbf{0.64} & -0.13
& 0.00 & \textbf{0.68} & 0.38
& 0.22 & \underline{0.63} & 0.40
& 0.28 & \underline{0.62} & 0.57 \\
GeLLM$^{3}$O
& 0.57 & \textbf{0.66} & 0.43
& 0.52 & \underline{0.59} & 0.41
& 0.44 & \underline{0.64} & 1.34
& 0.38 & \textbf{0.64} & 0.45
& 0.38 & \textbf{0.67} & 0.74 \\

\addlinespace
\multicolumn{16}{c}{\textbf{Agents with full MCP tools}} \\
Kimi-K3 Low (CC)
& \textbf{0.88} & 0.41 & 1.14
& 0.79 & 0.47 & 0.63
& 0.78 & 0.34 & \underline{4.83}
& 0.75 & 0.40 & 0.93
& \underline{0.82} & 0.37 & \textbf{2.73} \\
Kimi-K3 Low (Codex)
& 0.67 & 0.55 & 0.91
& \textbf{0.90} & 0.44 & \textbf{0.75}
& 0.71 & 0.37 & 4.52
& 0.68 & 0.43 & 0.86
& 0.72 & 0.41 & \underline{2.56} \\

\addlinespace
\multicolumn{16}{c}{\textbf{Ours}} \\
Kimi-K3 Low (R1)
& 0.63 & 0.57 & 0.76
& 0.66 & 0.46 & 0.29
& 0.72 & 0.45 & 2.97
& 0.76 & 0.46 & 0.89
& 0.59 & 0.52 & 1.50 \\
Kimi-K3 Low (R2)
& 0.74 & 0.49 & 0.92
& 0.75 & 0.41 & 0.54
& \underline{0.84} & 0.38 & 3.99
& \underline{0.79} & 0.37 & \underline{1.01}
& 0.70 & 0.46 & 2.17 \\
Kimi-K3 Low (R3)
& \underline{0.84} & 0.43 & 1.16
& 0.79 & 0.38 & \underline{0.68}
& \textbf{0.94} & 0.33 & 4.80
& \textbf{0.83} & 0.34 & \textbf{1.10}
& \textbf{0.83} & 0.44 & 2.55 \\
\bottomrule
\end{tabular}%
}
\end{table*}




\begin{table*}[t]
\caption{SMDD-Bench-93 task success rates. Single solution evaluates the final molecule; graph-any tests whether any recorded candidate satisfies all task constraints and Nesso activity criteria. SR is on a 0--1 scale. Bold/underline mark the best/second-best values.}
\label{tab:smdd_bench93_success}
\centering
\begin{tabular}{@{}lcc@{}}
\toprule
Model
& Single solution SR$\uparrow$
& Graph-any SR$\uparrow$ \\
\midrule

Kimi-K3 Low (Claude Code)
& 0.42 & 0.47 \\
Kimi-K3 Low (Codex)
& \underline{0.48} & \underline{0.57} \\

Kimi-K3 Low (M3OS R3)
& \textbf{0.59} & \textbf{0.61} \\
\bottomrule
\end{tabular}
\end{table*}

\subsection{Ablation Study}\label{sec:ablation}

\subsubsection{Generator composition and reasoning effort.}

Combining the Creative and Rational branches yields the highest third-round joint success rate among the low-effort generator configurations on MuMOInstruct-100: 0.91 for complete M3OS, compared with 0.74 for Creative-only and 0.70 for Rational-only (Fig.~\ref{fig:ablation}a). The complete system progresses from 0.67 at R1 to 0.91 at R3, while Creative-only and Rational-only progress from 0.72 to 0.74 and from 0.58 to 0.70, respectively. The complete configuration's advantage lies in joint success: a larger fraction of its selected candidates improves every requested property simultaneously. The configurations retain distinct property-improvement and structural-similarity profiles, with third-round RI of 2.24 for Creative-only, similarity of 0.42 for Rational-only, and RI/similarity of 2.19/0.37 for complete M3OS.

Higher reasoning effort further improves third-round performance, increasing joint SR from 0.91 to 0.94 and RI from 2.19 to 2.40 (Fig.~\ref{fig:ablation}a). Mean similarity is 0.37 at low reasoning effort and 0.35 at high reasoning effort.


\subsubsection{Retrieved context.}
On the 30-record hERG/BBBP subset, Rational-only generation with both retrieval sources achieves first-round loose/strict success of 0.86/0.60, compared with 0.78/0.54 without case retrieval and 0.84/0.57 without medicinal-chemistry retrieval (Fig.~\ref{fig:ablation}b). At R3, the corresponding rates are 0.94/0.80, 0.91/0.76 and 0.94/0.79. Across these two tasks, the observed success-rate differences are larger when case retrieval is removed.

\subsubsection{Search depth.}
\begin{figure}[t]
\centering
\includegraphics[width=\linewidth]{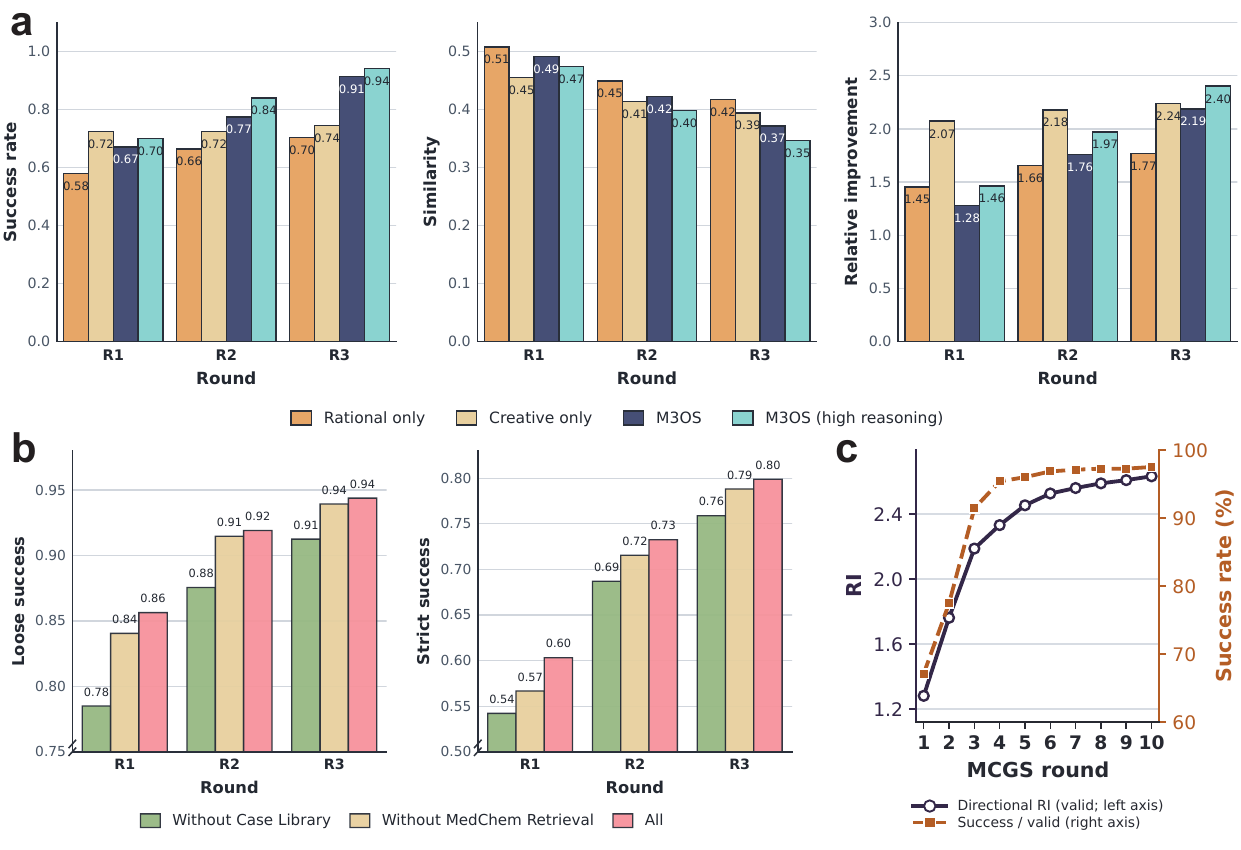}
\caption{Ablation study. \textbf{a}, MuMOInstruct SR, similarity and RI for Rational-only, Creative-only and complete M3OS at low reasoning effort, plus complete high-effort M3OS. SR: success rate; RI: relative improvement. \textbf{b}, Loose/strict success on BBBP/hERG (15 records each), all using Rational-only generation; ``All'' retains both retrieval sources. \textbf{c}, Ten-round complete low-effort M3OS: RI on the left axis, SR (\%) on the right. Panels use cumulative oracle top-20 selection and equal task averaging; MuMOInstruct conditions on valid candidates.}
\label{fig:ablation}
\end{figure}

M3OS reaches a joint SR of 96.0\% within five search rounds and 97.5\% within ten rounds on MuMOInstruct (Fig.~\ref{fig:ablation}c). SR increases from 67.1\% at R1 to 91.4\% at R3, while RI rises from 1.28 to 2.19, 2.45 and 2.63 at R1, R3, R5 and R10, respectively. The trajectories characterize how joint success and property improvement change as the cumulative search graph expands.

\subsection{Case Study}\label{sec:case_study}

The DHFR\_16 task from SMDD-Bench illustrates molecular optimization under coupled activity, structural and ADMET requirements. It requires a reduction of at least $0.30$ in the Nesso-predicted activity score, from $0.4098$ to $\leq 0.1098$, while maintaining Morgan-fingerprint Tanimoto similarity $T_c \geq 0.70$ and satisfying four ADMET windows and thirteen hard constraints. ADMET properties were evaluated with ADMET-AI \cite{swanson2024admetai}.

\begin{figure}[!htbp]
    \centering
    \includegraphics[width=\linewidth]{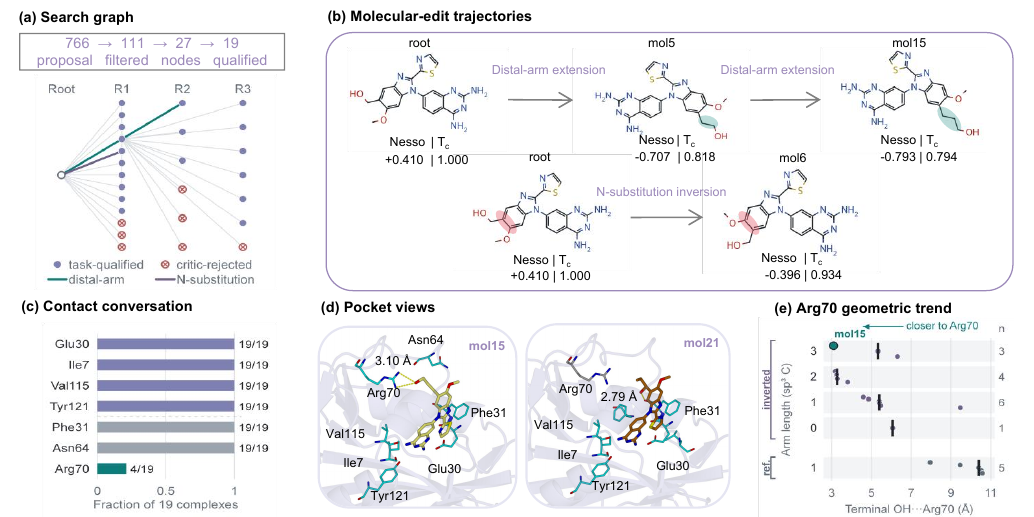}
    \caption{\textbf{Constrained molecular optimization on hDHFR.} \textbf{(a)} Search graph and candidate counts for three optimization rounds. Purple circles denote task-qualified candidates; crossed red circles denote critic-rejected candidates. Highlighted edges trace the molecular-edit paths shown in \textbf{(b)}.
    \textbf{(b)} Representative molecular-edit trajectories, with modified regions highlighted and search-time Nesso scores and Tanimoto similarities shown below each structure.
    \textbf{(c)} Fraction of qualified complexes contacting each residue; labels give counts.
    \textbf{(d)} Predicted pocket views of mol15 and mol21. Ligands are yellow and orange, respectively, and selected pocket residues are cyan. Annotated heavy-atom distances connect the terminal hydroxyl oxygen to Arg70 $N_{\eta2}$ in mol15 and to the Phe31 backbone carbonyl oxygen in mol21.
    \textbf{(e)} Terminal hydroxyl--Arg70 distance grouped by distal-arm length and $N$-substitution series. Points represent molecules, vertical ticks indicate row medians, and right-hand labels give sample sizes. Residues use mature hDHFR numbering.}
    \label{fig:case-study}
\end{figure}

Within three search rounds, M3OS identified \num{19} candidates satisfying the complete task contract (Fig.~\ref{fig:case-study}a). The graph links these candidates to their parent transformations, including two representative paths involving distal hydroxyl-arm extension and a change in benzimidazole $N$-substitution connectivity that preserves the molecular formula (Fig.~\ref{fig:case-study}b). Across the run, \num{766} proposals yielded \num{111} structures passing the basic filters and \num{26} evaluated graph candidates, alongside the reference root.

The diaminoquinazoline anchor remained unchanged across all \num{19} qualified molecules. In their Boltz-2-predicted hDHFR complexes \cite{passaro2025boltz}, contacts to Glu30, Ile7, Val115, Tyr121, Phe31 and Asn64 were retained throughout the population, whereas Arg70 contacts occurred in four complexes (Fig.~\ref{fig:case-study}c). The representative pocket views show distinct distal geometries: mol15 places its terminal hydroxyl oxygen \SI{3.10}{\angstrom} from Arg70 $N_{\eta2}$, while mol21 places its terminal hydroxyl oxygen \SI{2.79}{\angstrom} from the Phe31 backbone carbonyl oxygen (Fig.~\ref{fig:case-study}d). Across the $N$-inverted series, longer arms were associated with shorter hydroxyl--Arg70 distances (Spearman $\rho=-0.47$, $n=14$; Fig.~\ref{fig:case-study}e). The two edit series preserve the diaminoquinazoline anchor while varying distal-arm geometry and benzimidazole $N$-substitution.

\section{Discussion}\label{sec:discussion}

M3OS makes the evaluated molecular candidate the shared unit of coordination across agents: proposals are associated with a specific parent, assessed against a common task contract and retained for subsequent expansion. This organization connects the local decision of which structural edit to propose with the longer-term decision of which molecular trajectory to continue. The generator comparison identifies joint property improvement as the clearest advantage of the complete configuration, while SMDD-Bench-93 tests whether the system delivers a final molecule satisfying the coupled design constraints. These results support the evaluated system configuration; isolating the effect of MCGS parent selection requires comparisons with alternative selection policies under a common budget.


 The current evaluation uses predicted properties, so prospective synthesis and assays are needed to establish the experimental activity and ADMET profiles of selected candidates. Prediction errors and heuristic Critic scores can influence subsequent search priorities, motivating uncertainty-aware evaluation and independent experimental feedback. Incorporating measured properties into the same molecular state would extend the workflow toward iterative design--make--test cycles. M3OS provides a concrete architecture for keeping molecular proposals, evaluations and search decisions connected throughout this process.

\section{Methods}\label{sec:methods}

\subsection{Task representation and shared search state}\label{sec:problem_def}

A task specifies an initial molecule $m_0$, a natural-language objective, explicit structural or property requirements, and optional protein context. The Chief Architect converts these inputs into a task brief and search budget. Relevant ADMET optimization directions are specified at initialization; supplied thresholds, intervals, baseline-relative objectives and hold tolerances become an executable task contract. Numerical eligibility thresholds are taken from the supplied task requirements; objectives specified by direction are represented as directional criteria.

The persistent directed graph $\mathcal{G}_t=(V_t,E_t)$ links candidate identity to design history. Nodes are identified by canonical SMILES and retain generation provenance, evaluations and search statistics; edges record parent--child transformations. The initial molecule is the root; evaluated intermediate structures remain available alongside selected final candidates. Rediscovered molecules reuse existing nodes with additional incoming transformation edges. Both generator branches receive completed-round screening history, including remaining target gaps and failure reasons, while maintaining separate reasoning memories. The graph supplies candidate-level evidence and search statistics for subsequent node selection, while preserving the transformation history for analysis (Appendix~\ref{app:mcgs}).

\subsection{Graph-guided optimization loop}\label{sec:mcgs}

After evaluating the initial molecule, MCGS ranks reachable nodes using an upper-confidence score computed from accumulated rewards, node visits and parent visits. An LLM selects the expansion base from this shortlist using task context and node-level evidence; the choice must identify a shortlisted, existing node. Appendix~\ref{app:mcgs} specifies the selection and update equations.

A shared analysis identifies conserved fragments, modifiable regions and optimization priorities. The Creative Molecule Explorer builds and screens a candidate pool, while the Rational Medicinal Designer proposes edits using molecular evidence, medicinal-chemistry knowledge and optimization cases. The branches run in parallel. Their proposals are structurally extracted by the Auditor, validated and deduplicated before common Critic evaluation against the task contract. The Critic integrates computational evidence to assess each candidate’s improvements and satisfaction of the task constraints. Parent-aligned structural comparisons accompany candidate representations.

The Critic combines tool records with medicinal-chemistry reasoning to assign a heuristic score $S_{\mathrm{critic}}(m)\in[0,1]$. Search uses
\begin{equation}
S_{\mathrm{graph}}(m)=\alpha S_{\mathrm{critic}}(m)+(1-\alpha)S_{\mathrm{sim}}(m,m_0),
\label{eq:graph_score}
\end{equation}
where $S_{\mathrm{sim}}$ is radius-2, 2,048-bit Morgan-fingerprint Tanimoto similarity to the initial molecule and $\alpha$ controls the weighting. Critic, similarity and graph scores remain separate from task-gate outcomes.

Molecularly valid, evaluated candidates enter the graph with their task-satisfaction status, preserving intermediate structures for further optimization. The highest graph score in each nonempty expansion supplies the reward propagated along the selected root-to-candidate path: visits increase by one and accumulated rewards by that score. Search ends at the requested expansion budget or task-specific stopping criteria, and final candidates are selected by graph score.

\subsection{Controlled updates and evidence interfaces}\label{sec:tools}

An extraction-only Auditor converts finalized generator and Critic responses into structured records, using only explicitly stated molecules and evidence. Graph updates require valid molecular representations and structured evaluations. Role-specific tool allowlists and local wrappers enforce execution restrictions, while procedural skills guide decisions (Appendix~\ref{app:tools}). Audited outputs, tool evidence and transformation records retain the provenance of candidate evaluations.

We organize medicinal-chemistry sources into an optimization-oriented document collection and a concept-oriented knowledge base (Fig.~\ref{fig:knowledge-case}a). The MedChem Retrieval Agent returns concise, source-linked guidance while retaining detailed retrieval context within the agent. This interface supplies focused evidence for design decisions; construction and retrieval procedures are detailed in Appendix~\ref{app:medchem_rag}.

We organize property-specific molecular optimization relations into directed graphs and retrieve scaffold-matched connected components for the Rational Medicinal Designer (Fig.~\ref{fig:knowledge-case}b,c). These subgraphs provide related transformation examples and model-generated rationales for candidate edits, connecting retrieved cases to the current molecular context. Graph construction, preprocessing and resource statistics are detailed in Appendix~\ref{app:molecular-optimization-graphs}.

On request, the Reporter converts a read-only graph snapshot into an HTML report subject to identity, coverage and consistency checks (Appendix~\ref{app:reporting}).

\backmatter



\section*{Competing interests}
The authors declare no competing interests.

\section*{Data availability}
The benchmark task identifiers, generated molecules, search-graph traces and evaluation scripts supporting the reported results are available at \url{https://drive.google.com/drive/folders/12m4yYNW5deKtYipsdaqmWrdZkbIr2Hi6?usp=drive_link}. 

\section*{Code availability}
The code of M3OS and MCP Server is freely available at \url{https://github.com/wwwangjunjie/M3OS} and \url{https://github.com/wwwangjunjie/M3OS_MCP}.

\section*{Acknowledgements}
This work was supported by the National Natural Science Foundation of China (Grant No. 82607511) and Eastern Talent Plan (QNZH2025122).

\begin{appendices}

\section{Tools and procedural instructions}\label{app:tools}

\subsection{Computational capabilities and execution boundaries}

The tool layer combines external computational services exposed through MCP with local functions that bind the current task and enforce role-specific restrictions. Table~\ref{tab:M3OS-key-tools} summarizes the computational capabilities. Generators propose and screen structures, the Critic evaluates the complete valid candidate set, and the MedChem Retrieval Agent mediates external medicinal-chemistry retrieval. The Auditor extracts information from finalized responses, and the Reporter consumes the graph handoff. Procedural instructions specify role-level decision rules, and interface allowlists define the operations available to each role.

\begin{table}[t]
\caption{Computational capabilities used by the current M3OS workflow.}
\label{tab:M3OS-key-tools}
\begin{tabular}{@{}p{0.30\linewidth}p{0.64\linewidth}@{}}
\toprule
Task & Primary Tool Components \\
\midrule
Molecular and protein data retrieval &
UniProt \cite{uniprot2015uniprot}, ChEMBL \cite{gaulton2012chembl}, STOUT \cite{rajan2024stout} \\

Medicinal-chemistry knowledge retrieval &
Curated MedChem Resources, Synto \cite{synto2026github} \\

Literature search and question answering &
DuckDuckGo, PubMed, Semantic Scholar, MinerU \cite{wang2024mineru} \\

SMILES processing, validation, and topological graph generation &
RDKit \cite{bento2020open} \\

Molecular generation &
REINVENT4 \cite{loeffler2024reinvent} with Mol2Mol \cite{tibo2024exhaustive}, LinkInvent \cite{guo2023link}, and LibInvent \cite{fialkova2021libinvent} \\

ADMET prediction and constraint screening &
ADMET-AI \cite{swanson2024admetai}; RDKit \\

Optimization case retrieval &
Custom-built Molecular optimization case library \\

Target-activity prediction and screening &
Nesso-1 \cite{shenoy2026nesso} \\

Complex structure and interaction analysis &
Boltz-2 \cite{passaro2025boltz} and PLIP \cite{salentin2015plip} \\
\bottomrule
\end{tabular}
\end{table}

\subsection{Procedural skills}\label{sec:skills}

Procedural skills guide recurring decisions during molecular optimization: selecting generation and screening strategies, assessing the applicability of retrieved transformations, applying common evaluation criteria, and organizing molecular evidence for search and reporting. Table~\ref{tab:skill-list} lists the skills and role assignments used by the implementation. 

\begin{table*}[!ht]
\caption{Active procedural skills and their role assignments in M3OS.}
\label{tab:skill-list}
\centering
\small
\begin{tabular}{@{}p{0.23\textwidth}p{0.17\textwidth}p{0.52\textwidth}@{}}
\toprule
Skill & Assigned agent(s) & Description \\
\midrule
Candidate-quality gating & Critic & Evaluates all current-round candidates with frozen ADMET endpoints or explicit task constraints, then combines quantitative evidence with medicinal-chemistry risk review. \\
Nesso activity evaluation & Critic & Evaluates both generator branches with Nesso, adds comparison baselines, and preserves deterministic activity gates, scores and ranks. \\
Case-guided medicinal design & Rational & Converts pharmacophore and SAR hypotheses, medicinal-chemistry knowledge and retrieved optimization cases into interpretable, goal-calibrated molecular edits. \\
Medicinal-chemistry knowledge retrieval & Retrieval & Routes focused questions through shared memory, curated sources, the concept-oriented knowledge base and, when required, external literature. \\
Molecular optimization case retrieval & Rational & Retrieves scaffold-similar subgraphs from property-specific evolution graphs and extracts transferable transformation signals with their applicability limits. \\
Concept-oriented knowledge retrieval & Retrieval & Selects concept, article or multi-article retrieval according to query scope. \\
Protein-context assessment & Rational & Interprets initial-complex PLIP contacts as evidence for conserved features and modification hypotheses, retaining uncertainty about predicted structures and unmodeled candidates. \\
REINVENT candidate generation and filtering & Creative & Selects LibInvent, LinkInvent or Mol2Mol generation and one task-appropriate ADMET, activity or constraint-screening route. \\
Lead-optimization intersection screening & Creative & Applies explicit hard, ADMET and hold constraints; for activity tasks, screens the complete qualified set with Nesso before final selection. \\
SMILES validity gating & Creative, Rational, Critic & Validates final SMILES in batch; generators may attempt up to three intent-preserving repairs, while the Critic may only correct copied structures against valid input candidates. \\
Optimization report synthesis & Reporter & Converts the versioned molecular-search-graph handoff and programmatically generated molecular assets into a self-contained HTML decision report while preserving complete graph coverage, molecular numbering and parent--child provenance. \\
\bottomrule
\end{tabular}
\end{table*}

\subsection{Callable interface inventory}

Table~\ref{tab:tool-list} lists the current M3OS orchestration functions, task-specific wrappers and supporting MCP domain interfaces. The callable set in a session depends on configured endpoints, successful discovery, disabled-tool settings and role allowlists. Local wrappers that reuse an MCP function name are represented by the corresponding service entry, with their restrictions described in Section~\ref{sec:tools}. The general molecular utilities and source-path resolver remain supporting interfaces even when absent from default reasoning-agent allowlists. Operational health checks, framework planning/filesystem utilities, broad wiki administration and verbatim-source wiki interfaces are excluded from this domain inventory. Routine wiki retrieval exposes only the five article/concept interfaces listed below.

\begingroup
\small
\setlength{\LTpre}{0pt}
\setlength{\LTpost}{0pt}
\begin{longtable}{@{}p{0.34\linewidth}p{0.62\linewidth}@{}}
\caption{M3OS orchestration functions and supporting MCP domain interfaces.}\label{tab:tool-list}\\
\toprule
Function & Description \\
\midrule
\endfirsthead
\multicolumn{2}{@{}l}{\tablename\ \thetable\ (continued)}\\
\toprule
Function & Description \\
\midrule
\endhead
\midrule
\multicolumn{2}{r@{}}{Continued on next page}\\
\endfoot
\bottomrule
\endlastfoot

\multicolumn{2}{@{}l}{\textit{Local M3OS orchestration}}\\*
\nolinkurl{get_uploaded_file_info} & Documents and metadata uploaded to the session. \\
\nolinkurl{ask_medchem_knowledge} & Concise, source-aware guidance from the MedChem Retrieval Agent. \\
\nolinkurl{run_mcgs_search} & Graph search with the requested objective, generator mode, expansion-round limit and candidate target. \\
\nolinkurl{generate_optimization_report} & Validated self-contained HTML decision report from the current graph and optimization trace. \\
\nolinkurl{screen_reinvent_candidates_by_constraints} & Hard/ADMET screening of the current generation CSV; returns at most five passing Creative candidates. \\
\nolinkurl{screen_reinvent_candidates_by_intersection} & Nesso activity gates over the qualified set; retains at most ten candidates with screening evidence. \\
\nolinkurl{evaluate_candidate_constraints} & Scores the full current-round candidate set against the task contract, including failures. \\
\nolinkurl{evaluate_nesso_activity} & Nesso prediction with initial and current ligand references; returns raw metrics, task gates, scores and ranks. \\
\addlinespace

\multicolumn{2}{@{}l}{\textit{General molecular and retrieval MCP server}}\\*
\nolinkurl{web_search}, \nolinkurl{get_protein_uniprot_ids}, \nolinkurl{get_protein_sequence}, \nolinkurl{get_molecule_smiles}, \nolinkurl{answer_research_paper_question}, \nolinkurl{download_research_papers} & Web, UniProt and ChEMBL lookup; retrieval-augmented question answering over open-access papers. \\
\nolinkurl{molecular_similarity_search}, \nolinkurl{validate_smiles_string}, \nolinkurl{validate_smiles_batch}, \nolinkurl{clean_smiles_dataset} & Morgan-fingerprint similarity and RDKit SMILES validation utilities. \\
\nolinkurl{smiles_to_fragments_str}, \nolinkurl{generate_molecule_topology_image}, \nolinkurl{generate_molecule_difference_image}, \nolinkurl{generate_molecule_pair_difference_image} & Functional-group decomposition (AccFG) and molecular depiction with difference highlighting. \\
\addlinespace

\multicolumn{2}{@{}l}{\textit{IUPAC naming MCP server}}\\*
\nolinkurl{generate_iupac_name} & IUPAC names for molecular SMILES by STOUT. \\
\addlinespace

\multicolumn{2}{@{}l}{\textit{ADMET-AI MCP server}}\\*
\nolinkurl{admet_filter_by_admetai}, \nolinkurl{admet_predict_by_admetai}, \nolinkurl{admet_filter_by_task_constraints} & ADMET endpoint prediction and explicit RDKit/ADMET constraint screening. \\
\addlinespace

\multicolumn{2}{@{}l}{\textit{Nesso activity MCP server}}\\*
\nolinkurl{nesso_filter_by_cofolding}, \nolinkurl{nesso_predict_by_cofolding} & Binding-probability and affinity prediction with cached protein--ligand results. \\
\addlinespace

\multicolumn{2}{@{}l}{\textit{REINVENT4 MCP server}}\\*
\nolinkurl{setup_generation_Mol2Mol_LinkInvent}, \nolinkurl{prepare_smi_input}, \nolinkurl{run_generation}, \nolinkurl{generate_similarity_constrained_mol2mol}, \nolinkurl{setup_libinvent}, \nolinkurl{prepare_scaffold} & Mol2Mol, LinkInvent and LibInvent configuration and generation runs. \\
\addlinespace

\multicolumn{2}{@{}l}{\textit{Medicinal-chemistry source-catalog MCP server}}\\*
\nolinkurl{list_medchem_sources}, \nolinkurl{resolve_medchem_source_paths}, \nolinkurl{read_medchem_sources} & Listing, path resolution and provenance-aware reading of curated MedChem sources. \\
\addlinespace

\multicolumn{2}{@{}l}{\textit{Molecular-evolution-case-graph MCP server}}\\*
\nolinkurl{query_evolutionary_optimization_cases}, \nolinkurl{extract_molecule_scaffold} & Scaffold-similar evolution-subgraph retrieval and Bemis--Murcko scaffold extraction. \\
\addlinespace

\multicolumn{2}{@{}l}{\textit{Boltz/PLIP MCP server}}\\*
\nolinkurl{generate_protein_ligand_complex}, \nolinkurl{analyze_protein_ligand_interactions} & Protein--ligand complex generation and PLIP contact extraction. \\
\addlinespace

\multicolumn{2}{@{}l}{\textit{M3OS wiki MCP server by Synto}}\\*
\nolinkurl{read_article}, \nolinkurl{find_concept}, \nolinkurl{search_articles}, \nolinkurl{get_concept}, \nolinkurl{answer_question} & Approved-article and concept retrieval over the concept-oriented knowledge base. \\

\end{longtable}
\endgroup

\section{Medicinal-Chemistry Knowledge Base Construction and Retrieval}\label{app:medchem_rag}

\begin{figure}[!ht]
\centering
\includegraphics[width=\linewidth]{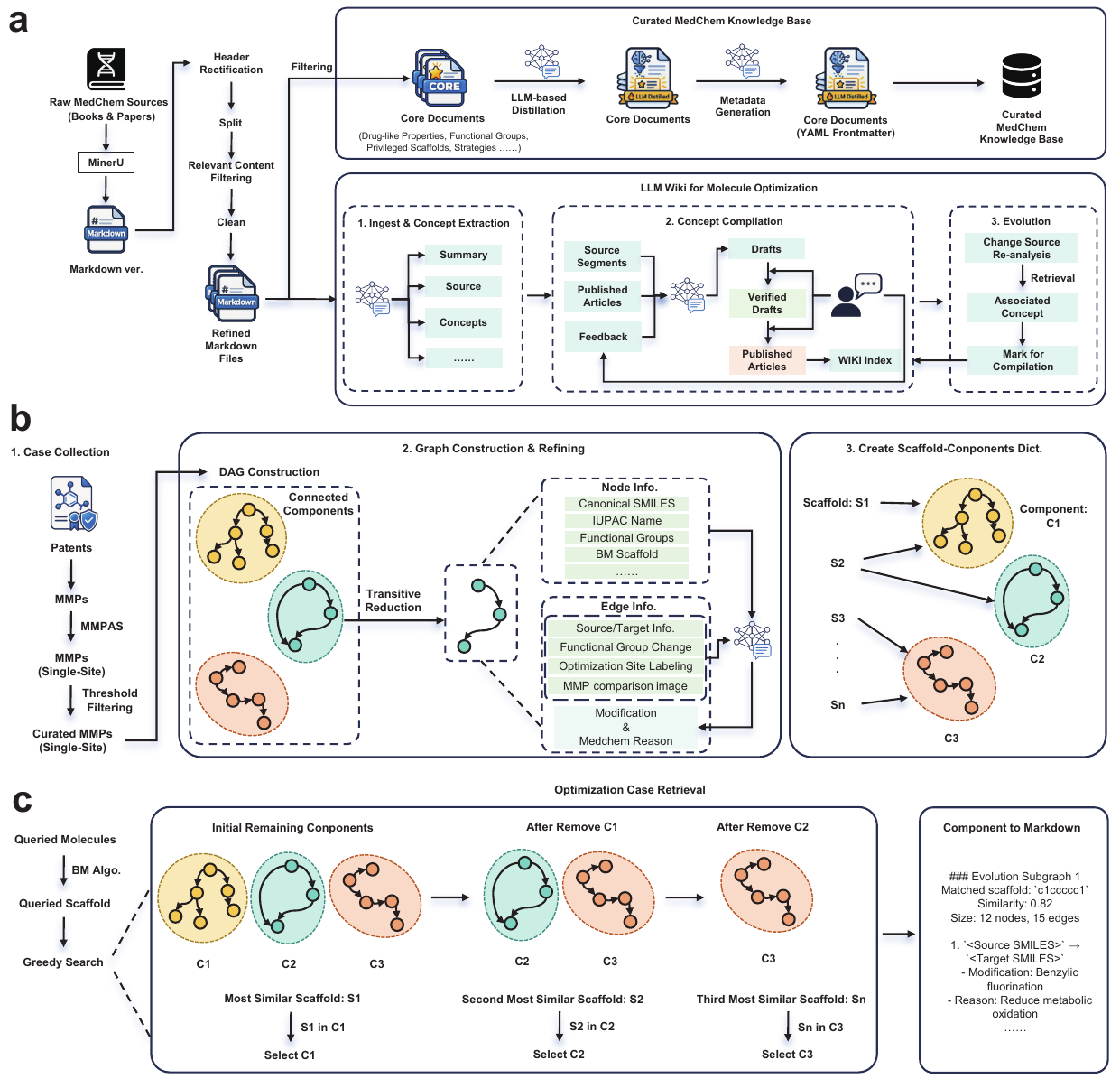}
\caption{Knowledge and case resources. \textbf{a}, Source documents form an optimization-oriented collection and concept wiki. \textbf{b}, Matched molecular pairs form annotated optimization graphs. \textbf{c}, Scaffold matching retrieves connected transformation examples for rational design. Case-graph construction is detailed in Appendix~\ref{app:molecular-optimization-graphs}.}
\label{fig:knowledge-case}
\end{figure}

Fig.~\ref{fig:knowledge-case}a summarizes knowledge construction; panels b,c show the case-graph construction and retrieval described in Appendix~\ref{app:molecular-optimization-graphs}.

The optimization-oriented knowledge collection was constructed from 1,221 documents covering medicinal chemistry, pharmacology, drug-design methodology and drug-development case studies, supplemented by 45 chapters describing functional groups and privileged scaffolds. An instruction-guided distillation procedure was applied to retain only information directly relevant to molecular optimization with instructions to exclude unrelated material and not introduce chemical claims unsupported by the source documents. This process yielded 86 core documents, comprising 16 documents on structural optimization strategies and empirical rules, eight on contemporary molecular-design tactics, 17 on optimization of drug-like properties, eight on functional groups and 37 on privileged scaffolds. Each distilled document was accompanied by a concise description of its scope, which was used during retrieval to identify relevant documents before introducing their full content into the model context.

The persistent concept-oriented knowledge base was implemented using Synto \cite{synto2026github} following the LLM Wiki framework \cite{karpathy2026llmwiki}. During source ingestion, each document was assigned a summary, language label and quality category. A language model then extracted a bounded set of candidate concepts, with the maximum number determined by source quality: up to eight concepts were retained from high-quality sources, four from medium-quality sources and two from low-quality sources. Candidate terms were reconciled against an existing registry of canonical concepts and aliases so that synonyms, abbreviations and spelling variants could be mapped to the same concept. Each source-level record retained its summary, quality assessment and associated canonical concepts, thereby preserving an explicit connection between the original source and the resulting knowledge representation.

Concept articles were generated by jointly considering all sources associated with a canonical concept. A higher-capacity language model integrated this evidence with the previous version of the article, when available, to produce a single synthesized representation. Articles contained links to related concepts and statement-level references to supporting sources. Their metadata included aliases, subject tags, contributing sources, source-quality information, confidence, review status and generation history. Newly generated or updated articles were assigned draft status and excluded from routine retrieval until approval. Automated structural quality control additionally identified incomplete metadata, unresolved inter-concept relationships, isolated concepts and weakly connected regions of the knowledge graph for subsequent review. At the time of analysis, the knowledge base contained 1,199 source documents, 1,197 source-level records, 829 approved concept articles and 202 articles awaiting review.

A dedicated MedChem Retrieval Agent mediated access to these resources. The agent first inspected relevant information retained in short-term memory to avoid redundant retrieval. When additional information was required, it queried both the optimization-oriented collection and the concept-oriented knowledge base. Focused questions concerning a specific chemical concept were resolved primarily against the corresponding concept article. When query terminology was ambiguous, a small set of candidate concepts was retrieved and compared before selecting the most relevant representation. Broader questions requiring information distributed across several topics were addressed by synthesizing evidence from multiple concept articles. Only approved concept-level articles were available during routine retrieval; draft articles, corpus-wide inventories and verbatim source passages were excluded from this retrieval pathway.

Public web search and research-paper retrieval supplement the curated resources when a question requires additional coverage, recent findings or primary evidence. Detailed documents remain within the MedChem Retrieval Agent, which returns a concise synthesis of task-relevant evidence, source references and applicability conditions.

\section{Construction of Molecular Evolutionary Graphs}\label{app:molecular-optimization-graphs}


Historical molecular-optimization cases were curated from proprietary patents and organized into property-specific directed graphs for BBBP, hERG liability, human liver microsomal stability, and apparent permeability ($P_{\mathrm{app}}$). For each property $p$, a graph $G_p=(V_p,E_p)$ was constructed, where nodes represent unique molecules and directed edges encode reported before–after optimization relationships. Each edge is directed toward the molecule with the improved target-property value. To derive these relationships, molecular pairs were extracted using the Matched Molecular Pair Analysis (MMPA) tool \cite{geng2023novo, griffen2011matched}, retaining only those involving a single-site change that conferred a measurable improvement in property value. Table~\ref{tab:mmp-graphs} summarizes the resulting resources, and Fig.~\ref{fig:knowledge-case}b,c illustrates the construction and retrieval workflow.

\begin{table}[t]
\caption{Property-specific molecular evolution graphs used by M3OS. Components are weakly connected components of the directed graph.}\label{tab:mmp-graphs}
\centering
\begin{tabular}{@{}lrrrrr@{}}
\toprule
Property & Nodes & Edges & Roots & Components & Largest component \\
\midrule
BBBP & 674 & 665 & 331 & 155 & 35 \\
hERG & 1,934 & 2,651 & 1,063 & 360 & 54 \\
Human liver microsomal stability & 2,391 & 3,956 & 1,228 & 388 & 80 \\
Apparent permeability ($P_{\mathrm{app}}$) & 2,171 & 5,349 & 1,019 & 293 & 163 \\
\botrule
\end{tabular}
\end{table}

Input SMILES were parsed with RDKit and converted to canonical isomeric SMILES. Invalid structures and molecular pairs that became identical after canonicalization were excluded. Duplicate directed relations were merged while retaining their evidence counts and annotations.

Each property-specific graph was constrained to be acyclic. Transitive reduction was then applied so that redundant direct edges were removed while preserving reachability; the corresponding evidence and replacement paths were stored separately.

Each node was annotated with a structure-preserving Bemis--Murcko scaffold \cite{bemis1996properties}. For every retained edge, the molecular pair was aligned using a maximum common substructure with complete ring matching and compatible valence states. The paired depiction highlighted conserved and modified structural features and was used by a vision-capable language model to generate a structural-transformation description and property-specific medicinal-chemistry rationale.

For retrieval, query and graph scaffolds were represented using radius-2, 2,048-bit Morgan fingerprints and compared by Tanimoto similarity \cite{morgan1965generation,bajusz2015tanimoto}. Weakly connected components were ranked by their best scaffold match to the query, and up to three distinct components were returned for each requested property.

The transformation descriptions and medicinal-chemistry rationales are model-generated annotations of the source-derived molecular relations. They serve as design hypotheses presented to the Rational Medicinal Designer rather than verified medicinal-chemistry claims; whether a suggested edit is adopted is determined by the subsequent tool-based candidate evaluation, not by the annotation itself.

\section{Optimization Report Generation and Validation}\label{app:reporting}

\begin{figure}[!ht]
\centering
\includegraphics[width=\linewidth]{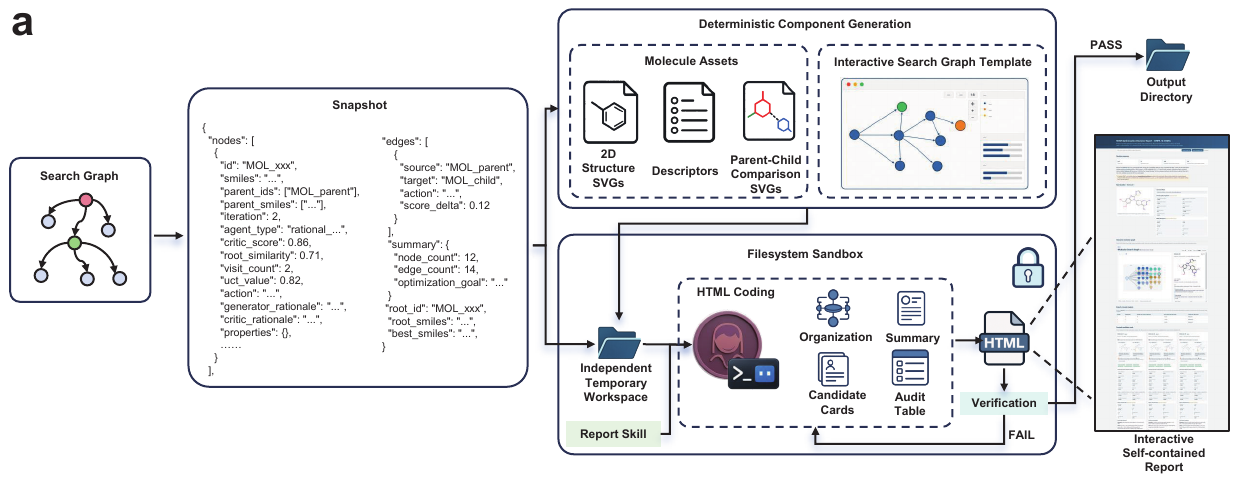}
\caption{Search-report generation. Molecular depictions, predictions, rationales and an interactive graph are assembled from a graph snapshot into an HTML report. Validation gates delivery; PASS/FAIL indicate check outcomes.}
\label{fig:reporting}
\end{figure}

\begin{figure}[!ht]
\centering
\includegraphics[width=\linewidth]{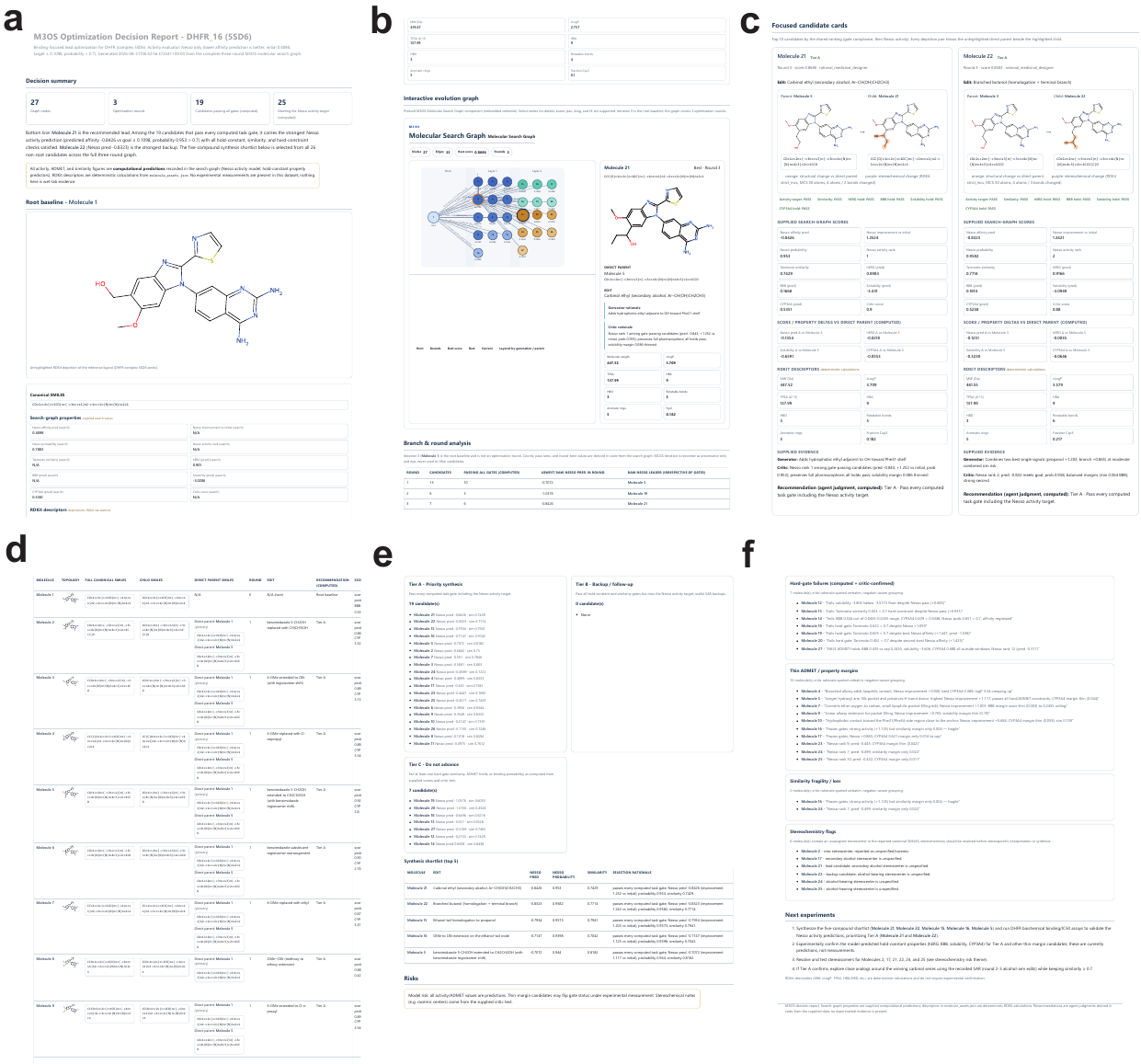}
\caption{Example of an M3OS optimization decision report. Selected views of the self-contained HTML report generated from the three-round search graph for the DHFR\_16 task. \textbf{a}, Decision summary and reference-molecule profile. \textbf{b}, Interactive molecular search graph with node-level details and round-wise statistics. \textbf{c}, Focused candidate cards showing direct-parent structural comparisons with highlighted edits, predicted activity and ADMET properties, task-constraint checks, and generator and Critic rationales. \textbf{d}, Excerpt from the full-node audit, linking molecule identifiers to structures, complete canonical SMILES and recorded parent--child relationships. \textbf{e}, Recommendation tiers and a five-compound synthesis shortlist. \textbf{f}, Risk summaries and proposed follow-up experiments. Activity and ADMET values are computational predictions, whereas RDKit descriptors are deterministic calculations; recommendations remain subject to experimental validation.}
\label{fig:optimization-report-example}
\end{figure}

Fig.~\ref{fig:reporting} summarizes the graph-to-report generation and validation workflow, while Fig.~\ref{fig:optimization-report-example} presents an example report for the DHFR\_16 task, illustrating how recorded optimization trajectories and candidate-level evidence are organized for post-search inspection.

\subsection{Graph snapshot and handoff.}
When a report is requested, M3OS deep-copies the current search-graph snapshot and, when the live graph is available, enriches it with parent-specific transformation metadata. The resulting versioned JSON handoff contains the task objective, complete node and edge sets, molecular representations, search scores and properties, structural edits, generator and Critic rationales, and report-coverage controls. The graph remains the source of candidate identities, optimization outcomes and design rationales in the report; generated molecular assets provide deterministic depictions and descriptors of those recorded structures.

\subsection{Molecular assets and graph rendering.}
Before invoking the coding agent, M3OS assigns each graph node a deterministic, report-wide molecule number. The Full-Node Audit provides the authoritative mapping from these numbers to topology diagrams and full canonical SMILES, while summaries, recommendations and synthesis lists refer primarily to molecule numbers. RDKit-based routines generate unhighlighted topology depictions, deterministic molecular descriptors and parent-aligned structure-difference SVGs. A separate renderer constructs the complete Molecular Search Graph component from all graph nodes and valid parent--child edges, supporting node selection, detail inspection, zooming, panning, dragging and view fitting.

\subsection{Isolated document assembly.}
The coding agent runs in a disposable workspace containing the versioned JSON handoff, molecular assets, the prebuilt search-graph component and a dedicated reporting skill. The project source tree and private development references are not exposed to the reporting task. The agent implements the document structure, styling and interactions and embeds the prebuilt graph component verbatim. The report contains a summary of the main findings, a root profile, branch- and round-level analysis, focused parent--child candidate cards, the complete node audit, recommendation tiers, a synthesis shortlist, risk analysis and proposed next experiments. Interactive features include molecule filtering, synchronized reviewer selections and notes, CSV export and re-export of a self-contained HTML document. All CSS, JavaScript, molecular SVGs and state required for review are embedded in the report, eliminating runtime dependencies on external web resources.

\subsection{Validation before delivery.}
The generated report is released only after deterministic validation. Checks cover document completeness, absence of external stylesheets and remote assets, and verbatim inclusion of the prebuilt graph component. The validator requires exactly one graph node and detail record for every molecule and one rendered edge for every valid graph relationship. It also checks stable molecule numbering, direct-parent comparisons, complete non-truncated SMILES, candidate-card and audit coverage, synthesis-list consistency and persistent review state. Failure of any required check prevents report delivery. 

\section{Multi-agent Monte Carlo Graph Search Implementation}\label{app:mcgs}

\subsection{Task Initialization}\label{sec:task_init}

An information-extraction module processes the user request and session context to identify the initial molecule, optimization objective, candidate-generation requirements, search budget and optional protein context. A task brief summarizes the objectives, constraints and permitted or prohibited modifications. For monotonic ADMET objectives, the workflow selects directly relevant endpoints from the available metadata and freezes their optimization directions for downstream agents. A structured task contract represents executable thresholds, intervals, baseline-relative objectives and hold tolerances when supplied. 

\subsection{Session state, memory and structured records}\label{app:state_records}

The centralized session state contains the optimization objective and constraints, task contract and protein context, molecular search graph, current proposals, evaluation results and role-specific memories. Molecule-associated records use canonical SMILES, and target-dependent predictions retain protein identity. Completed-round screening records accumulate qualified-set metrics, target gaps and failure reasons for reuse by both generator branches. 

Graph nodes contain molecular representations, generation provenance, property and evaluation records, and search statistics. Parent--child edges contain transformation and generation metadata, including alternative incoming transformations when a molecule is rediscovered. The extraction-only Auditor maps explicitly stated generator proposals and Critic results into predefined records without introducing additional molecules or evidence. Persistent records retain audited outputs, tool-grounded evidence, validation outcomes, graph changes and recovery events.

SMILES validation uses RDKit parsing and canonicalization. Generators may attempt up to three intent-preserving repairs of their final molecular representations. The Critic may only correct a copied structure against a valid input candidate. Failed representations are excluded before evaluation; failures in evaluation or structured extraction do not produce a valid candidate record for graph insertion. An evaluated candidate that fails task requirements may still be recorded in the graph, keeping record validity separate from task feasibility.

\subsection{Search Lifecycle}\label{sec:lifecycle}

\subsubsection{Overview.}\label{sec:overview}
The canonical initial molecule initializes the graph as its root, and an initial Critic evaluation establishes the graph state. Each subsequent iteration selects an expansion base, generates and evaluates candidates, incorporates valid results into the graph and updates the search statistics.

\subsubsection{Node selection.}\label{sec:node_select}
We adapt the MCGS node-selection rule to the persistent molecular search graph in M3OS. Equations (\ref{eq:exploit_value})–(\ref{eq:search_selection}) specify this system-specific formulation, which retains graph-based state reuse, accumulated reward and visit statistics, and UCB-guided candidate ranking.
For a node $v$ with accumulated reward $R_v$ and visit count $n_v$, the exploitation value is
\begin{equation}
Q_v=
\begin{cases}
R_v/n_v, & n_v>0,\\
R_v, & n_v=0.
\end{cases}
\label{eq:exploit_value}
\end{equation}
The corresponding parent-based visit statistic is
\begin{equation}
N_v=
\begin{cases}
\max\left(1,\sum_{u\in\mathrm{Parent}(v)}n_u\right), & v\ \text{is non-root},\\
\max(1,n_v), & v\ \text{is root}.
\end{cases}
\label{eq:parent_visits}
\end{equation}
A molecular node may have multiple parents because the same molecule can be reached through different transformations. We therefore sum the visit counts of its direct parents to represent the aggregate search activity in its predecessor neighborhood, providing a single node-level reference for exploration without choosing an arbitrary incoming parent. This parent-aggregated statistic is a heuristic adaptation for our molecular search graph. The exploration term still depends on the node-specific denominator \(\max(1,n_v)\), so sharing the same parent statistic does not necessarily produce identical exploration bonuses. The adapted UCB score is
\begin{equation}
U_v=Q_v+c\sqrt{\frac{\ln N_v}{\max(1,n_v)}},
\label{eq:search_selection}
\end{equation}
where $c$ controls exploration strength. Before each expansion, M3OS updates these scores for all reachable nodes and ranks them accordingly. An LLM-based selection module chooses the expansion base from the highest-ranked nodes using task context and node-level evidence. The selection is accepted only if it identifies both a shortlisted candidate and an existing graph node.

\subsubsection{Parallel molecular expansion.}\label{sec:expansion}
A shared analysis of the selected parent identifies conserved fragments, modifiable regions, structural risks and optimization priorities. The Creative Molecule Explorer constructs and screens a REINVENT candidate pool, while the Rational Medicinal Designer proposes modifications using medicinal-chemistry knowledge and retrieved optimization cases. Both agents receive the shared analysis and accumulated screening history from completed rounds, while retaining separate role-specific memories. The Auditor converts each finalized generator output into structured molecular proposals, which are then aggregated for validation.

\subsubsection{Candidate validation and Critic evaluation.}\label{sec:simulation}
RDKit parses and canonicalizes the generated candidates; invalid representations and duplicate canonical structures are removed before evaluation. Each valid candidate is visually compared with its parent using maximum-common-substructure alignment when available, highlighting conserved and modified regions and stereochemical changes. The depiction is associated with the canonical SMILES and supplied to the Critic alongside the symbolic representation.

The Critic submits the complete valid candidate set from both branches to the applicable evaluation tools. Monotonic ADMET tasks use the frozen endpoint policy, tasks with executable contracts use deterministic constraint evaluation, and target-activity tasks additionally use the Nesso wrapper with initial and current ligand references. Backends reuse cached predictions, and prior Creative screening does not exclude candidates from Critic evaluation. The Critic combines these records with comparative medicinal-chemistry reasoning to return a candidate-specific rationale, property record and heuristic score $S_{\mathrm{critic}}(m)\in[0,1]$. The deterministic activity rank remains a separate tool output. The Auditor converts the finalized response into structured evaluation records.

\subsubsection{Graph insertion and reward propagation.}\label{sec:propagation}
The graph search score combines Critic evaluation and similarity to the initial molecule $m_0$ as defined in Eq.~\eqref{eq:graph_score},
where $\alpha$ controls their relative contributions. Critic and graph scores are retained separately. The search implementation computes $S_{\mathrm{sim}}$ with radius-2, 2,048-bit Morgan bit vectors and rounds the Tanimoto coefficient to four decimal places. If similarity computation fails, its contribution to the graph score is zero.
Candidates with valid molecular representations and structured evaluations are incorporated into the graph, including evaluated candidates that fail task objectives or constraints. An unseen molecule creates a new node connected to the selected parent; an existing molecule receives an additional incoming transformation edge.

Let $\mathcal{B}_t$ denote the candidates generated in expansion $t$ with valid molecular representations and structured Critic evaluations. For a nonempty $\mathcal{B}_t$, the highest-scoring candidate and its reward are
\begin{equation}
v_t^{*}=\arg\max_{v\in\mathcal{B}_t}S_{\mathrm{graph}}(v),
\qquad
r_t=S_{\mathrm{graph}}(v_t^{*}).
\label{eq:expansion_reward}
\end{equation}
The reward is propagated along the selected root-to-candidate path. Each node $v$ on that path is updated as
\begin{equation}
n_v\leftarrow n_v+1,
\qquad
R_v\leftarrow R_v+r_t.
\label{eq:reward_propagation}
\end{equation}
These statistics inform upper-confidence-based selection in subsequent iterations.

\subsubsection{Termination and final selection.}\label{sec:terminal}
Search terminates when the requested expansion budget is reached or task-specific stopping criteria are satisfied. The final candidate set is then selected according to graph search scores. 

\subsection{Benchmark instruction content}\label{app:benchmark_prompts}

Benchmark inputs comprise the starting molecular structure and the requested property directions or task constraints. MolOpt-120 uses eight single- or double-property tasks; MuMOInstruct-100 uses five combinations of three or four properties; SMDD-Bench-93 additionally provides the protein target and coupled lead-optimization requirements. Candidate-output protocols differ by method class and are specified in Section~\ref{sec:metrics}: fixed-pool baselines generate 20 candidates per MolOpt or MuMOInstruct record, coding agents return up to 20, and M3OS is evaluated on a selected pool from its cumulative graph.

Within M3OS, the task brief carries objectives and permitted or prohibited modifications. Generator inputs include the selected parent, shared molecular analysis and prior screening feedback. Critic inputs comprise validated candidate structures, parent comparisons and tool evaluation records. Retrieval requests are focused knowledge questions; Auditor inputs are finalized generator or Critic responses. 

\subsection{Baseline prompting}\label{app:baseline_prompts}

Baseline prompts convey the starting molecule and optimization objectives through interfaces appropriate to each model or agent. Prompt wording, conversational structure and response conventions are adapted to the respective baseline.

\paragraph{Chemistry-specific language models.}
Prompts follow the styles and input conventions of the respective original works. DrugAssist uses natural-language molecule-editing questions, while GeLLM$^{3}$O retains its input, property-adjustment and response structure. ChemLLM and LlaSMol use their model-specific chat or instruction templates, and mCLM follows its molecule-generation input and output conventions. The benchmark molecule and requested property changes are incorporated into these formats. When a benchmark objective extends beyond a model's original task set, the instruction is adapted within that model's prompt style.

\paragraph{General-purpose language models.}
Direct LLM baselines use a common medicinal-chemistry expert system prompt and a shared task template containing the initial SMILES, optimization goal, requested candidate count and any additional task requirements. The prompt asks for chemically valid, distinct candidates and a structured response containing each candidate's SMILES, modification type, concise rationale and confidence score. This is a direct generation setting without access to external molecular-evaluation tools.

\paragraph{General-purpose coding agents.}
Codex and Claude Code receive the benchmark task together with instructions for tool use, candidate screening and final submission. Available protein and ligand structure files are identified when relevant to the task. The instructions include guidance to generate a candidate pool, apply inexpensive local filters, batch property predictions and prioritize candidates for activity evaluation within the stated tool-call budgets (Appendix~\ref{app:baselines}). The agents plan and execute tool calls within these instructions and can use returned evidence to refine their candidates. Final outputs comprise a ranked list of candidate SMILES and a designated best molecule. The shared task requirements are accompanied by the instructions and domain skills available in each agent runtime.

\section{Evaluation and run configuration}\label{app:exp}\label{sec:baselines}\label{app:baselines}\label{app:ablation_settings}

\subsection{Benchmark subset construction}\label{sec:datasets}\label{app:datasets}

For MolOpt-120, starting molecules were sampled at random without replacement from the 500-molecule MolOpt-Instructions test pool \cite{ye2025drugassist} and assigned to nonoverlapping task groups. The original iDrug endpoint was unavailable, so the reported solubility, BBBP and hERG evaluations use surrogate predictions.

For MuMOInstruct-100, records were sampled without replacement within each task from the official test split \cite{dey2025gellm3o}. Sampling at the record level yielded 96 unique starting molecules. Of the task combinations in Table~\ref{tab:mumo-baseline-results}, BBBP/pLogP/QED is in-domain and the remaining four are out-of-domain under the source benchmark's definitions.


\subsection{Search settings, model configuration, and token usage}
\label{app:search_settings_tokens}

\subsubsection{Search settings.}
The Monte Carlo graph search used a UCB exploration constant of $c=0.01$.
For each evaluated molecule $m$, the Critic--similarity
weight was $\alpha=0.8$. At each expansion, the 5 highest-UCB
eligible non-root nodes formed the shortlist from which an LLM selected
one expansion base. This search shortlist is distinct from the
20-candidate pool used for benchmark evaluation. A newly inserted node
started with zero visits and an accumulated reward equal to its initial
graph score, $n_v^{(0)}=0$ and
$R_v^{(0)}=S_{\mathrm{graph}}(v)$. The root instead started with the
score assigned by its initial Critic evaluation. No additional
backpropagated reward was assigned at initialization.

\subsubsection{Model and decoding configuration.}
M3OS and the two coding-agent baselines used the Kimi API model
identifier \texttt{kimi-k3} with
\texttt{reasoning\_effort=low}. Kimi K3 always operates in thinking
mode. The requests did not explicitly set temperature, nucleus-sampling
probability, sampling seed, or a completion-token cap. The official
Kimi-K3 API documentation specifies fixed values of
$\mathrm{temperature}=1.0$, $\mathrm{top\_p}=0.95$, $n=1$,
$\mathrm{presence\_penalty}=0$, and
$\mathrm{frequency\_penalty}=0$; these parameters are omitted from
requests. The documentation currently lists 131,072 as the default
\texttt{max\_completion\_tokens}, but this was not an explicitly
enforced limit in our runs. The API identifier was recorded, whereas
an immutable served-weight revision was not exposed in the experiment
logs. The generic benchmark parser's
\texttt{temperature=0.0} default was not forwarded to Kimi K3.

\subsubsection{Token accounting.}
Table~\ref{tab:appendix_token_usage} summarizes the aggregate token
consumption of Claude Code, Codex, and M3OS on the three benchmarks.
All values are reported in millions of tokens. Output
counts include the tokens used for internal thinking or reasoning in
addition to the final visible responses.

\begin{table}[t]
\caption{Aggregate token consumption across the three benchmarks.
All values are in millions of tokens.}
\label{tab:appendix_token_usage}

\centering
\footnotesize
\setlength{\tabcolsep}{3pt}

\begin{tabular*}{\linewidth}{@{\extracolsep{\fill}}llrr@{}}
\toprule
Benchmark & Harness & Input & Output \\
\midrule

MolOpt-120 & Claude Code
& 61.03 & 1.33 \\
MolOpt-120 & Codex
& 123.91 & 1.12 \\
MolOpt-120 & M3OS
& 63.46 & 2.71 \\
\addlinespace

MuMOInstruct-100 & Claude Code
& 57.20 & 1.31 \\
MuMOInstruct-100 & Codex
& 169.43 & 1.17 \\
MuMOInstruct-100 & M3OS
& 60.27 & 2.40 \\
\addlinespace

SMDD-Bench-93 & Claude Code
& 163.14 & 2.44 \\
SMDD-Bench-93 & Codex
& 747.52 & 3.51 \\
SMDD-Bench-93 & M3OS
& 224.21 & 2.47 \\
\midrule

All three & Claude Code
& 281.37 & 5.09 \\
All three & Codex
& 1040.86 & 5.80 \\
All three & M3OS
& 347.94 & 7.58 \\
\bottomrule
\end{tabular*}
\end{table}

These totals characterize the computational footprint of each complete
harness rather than the efficiency of the underlying language model in
isolation. The three harnesses differ in their interaction patterns,
numbers of model turns, retry behavior, context construction, and use
of prompt caching. In particular, the substantially larger Codex input
count on SMDD-Bench-93 reflects longer agent trajectories and repeated
attempts.

\subsection{Run configuration details}

For SMDD-Bench-93, each method is limited to eight Nesso batch submissions and 15 ADMET-AI batch submissions per instance. Each submission counts as one call regardless of the number of molecules in the batch.

The retrieval ablations in Section~\ref{sec:ablation} use Kimi-K3 Low. The extended-search runs retain all knowledge components.

\end{appendices}

\bibliography{sn-bibliography}

\end{document}